\documentclass{article} 
\usepackage{iclr2025_conference,times}

\usepackage{hyperref}
\usepackage{url}
\usepackage{microtype}
\usepackage{graphicx}
\usepackage{subcaption}
\usepackage{booktabs} 
\usepackage{makecell}
\usepackage{multirow}
\usepackage{tablefootnote}
\usepackage{pdflscape}
\usepackage{enumitem}
\usepackage{pifont}
\usepackage{lipsum}
\usepackage{wrapfig}
\usepackage{colortbl}
\usepackage{xcolor}

\usepackage{amsmath}
\usepackage{amssymb}
\usepackage{mathtools}
\usepackage{amsthm}
\usepackage[bb=dsserif]{mathalpha}

\usepackage{amsmath,amsfonts,bm}

\def\eqref#1{equation~\ref{#1}}

\def\1{\bm{1}}

\def\vtheta{{\bm{\theta}}}

\def\vx{{\bm{x}}}

\def\mC{{\bm{C}}}

\def\mW{{\bm{W}}}

\DeclareMathAlphabet{\mathsfit}{\encodingdefault}{\sfdefault}{m}{sl}
\SetMathAlphabet{\mathsfit}{bold}{\encodingdefault}{\sfdefault}{bx}{n}

\def\moirai{\textsc{Moirai}}
\def\moiraimoe{\textsc{Moirai-MoE}}
\def\moiraismall{{\moirai}\textsubscript{S}}
\def\moiraibase{{\moirai}\textsubscript{B}}
\def\moirailarge{{\moirai}\textsubscript{L}}
\def\moiraimoesmall{{\moiraimoe}\textsubscript{S}}
\def\moiraimoebase{{\moiraimoe}\textsubscript{B}}
\def\moiraimoelarge{{\moiraimoe}\textsubscript{L}}

\title{Moirai-MoE: Empowering Time Series Foundation Models with Sparse Mixture of Experts}

\author{Xu Liu$^{1,2}$\thanks{Work done during internship/industrial PhD at Salesforce AI Research.}, \ Juncheng Liu$^{1}$, Gerald Woo$^{1*}$, Taha Aksu$^{1}$, Yuxuan Liang$^{3}$, Roger Zimmermann$^{2}$,\\
\textbf{Chenghao Liu$^{1}$\thanks{Corresponding author. Email: chenghao.liu@salesforce.com}, \ Silvio Savarese$^{1}$, Caiming Xiong$^{1}$, Doyen Sahoo$^{1}$} \\
$^{1}$Salesforce AI Research, $^{2}$National University of Singapore, \\
$^{3}$The Hong Kong University of Science and Technology (Guangzhou) \\
}

\iclrfinalcopy 
\begin{document}

\maketitle

\begin{abstract}
Time series foundation models have demonstrated impressive performance as zero-shot forecasters, i.e., they can tackle a wide variety of downstream forecasting tasks without explicit task-specific training. However, achieving effectively unified training on time series remains an open challenge. Existing approaches introduce some level of model specialization to account for the highly heterogeneous nature of time series data. For instance, $\moirai$ pursues unified training by employing multiple input/output projection layers, each tailored to handle time series at a specific frequency. Similarly, TimesFM maintains a frequency embedding dictionary for this purpose. We identify two major drawbacks to this human-imposed frequency-level model specialization: (1) Frequency is not a reliable indicator of the underlying patterns in time series. For example, time series with different frequencies can display similar patterns, while those with the same frequency may exhibit varied patterns. (2) Non-stationarity is an inherent property of real-world time series, leading to varied distributions even within a short context window of a single time series. Frequency-level specialization is too coarse-grained to capture this level of diversity. To address these limitations, this paper introduces $\moiraimoe$, using a single input/output projection layer while delegating the modeling of diverse time series patterns to the sparse mixture of experts (MoE) within Transformers. With these designs, $\moiraimoe$ reduces reliance on human-defined heuristics and enables automatic token-level specialization. Extensive experiments on 39 datasets demonstrate the superiority of $\moiraimoe$ over existing foundation models in both in-distribution and zero-shot scenarios. Furthermore, this study conducts comprehensive model analyses to explore the inner workings of time series MoE foundation models and provides valuable insights for future research.
\end{abstract}

\section{Introduction}
Foundation models have transformed several fields, such as natural language processing \citep{llama3} and computer vision \citep{segmentanything}, demonstrating impressive zero-shot performance. Inspired by these successes, time series forecasting is experiencing a similar shift \citep{liang2024foundation}. The traditional approach of developing separate models for each dataset is being replaced by the concept of universal forecasting \citep{moirai}, where a pretrained model can be applied across diverse downstream tasks in a zero-shot manner, regardless of variations in domain, frequency, dimensionality, context, or prediction length. This new paradigm significantly reduces the complexity of building numerous specialized models, paving the way for forecasting-as-a-service.

\begin{figure}[t]
    \centering
    \includegraphics[width=0.97\textwidth]{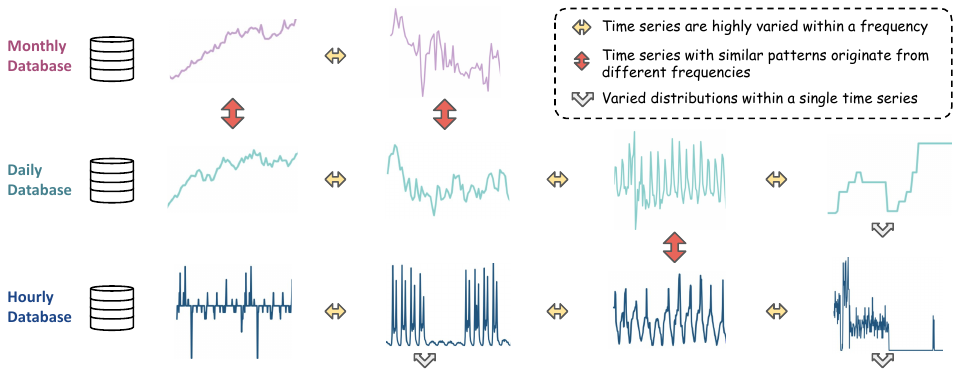}
    \caption{An illustration of the challenges arising from grouping time series by frequency and imposing frequency-level model specialization: the diversity of patterns within the same frequency group, the similarity of patterns across different frequencies, and the variability of distributions within a single time series. The examples presented are derived from \textbf{real time series} in the Monash benchmark \citep{monash}.}
    \vspace{-1em}
    \label{fig:intro}
\end{figure}

To excel in zero-shot forecasting, time series foundation models are pretrained on massive data from a variety of sources. However, unlike language and vision modalities which benefit from standardized input formats, time series data is inherently heterogeneous, posing significant challenges for \textit{unified time series training}. Existing solutions such as TEMPO \citep{tempo} and UniTime \citep{unitime} leverage language prompts to provide data identification information, thereby discerning the source of data and achieving model specialization at the dataset level. $\moirai$ \citep{moirai} goes a step further and proposes a more granular categorization based on a time series meta feature -- frequency. Specifically, they design multiple input/output projection layers with each layer specialized to handle data corresponding to a specific frequency, thereby enabling frequency-level specialization. Similarly, TimesFM \citep{timesfm} is also at this level of specialization, distinguishing the data by maintaining a frequency embedding mapping.

Given the heterogeneity of time series, we acknowledge the value of model specialization; however, we argue that \textit{human-imposed frequency-level specialization lacks generalizability and introduces several limitations}. (1) Frequency is not always a reliable indicator and might not effectively capture the true structure of time series data. As shown in Figure \ref{fig:intro}, time series with different frequencies can exhibit similar patterns, while those with the same frequency may display diverse and unrelated patterns. This human-imposed mismatch between frequency and pattern undermines the efficacy of model specialization, resulting in inferior performance. (2) Furthermore, real-world time series are inherently non-stationary \citep{nonstationary}, displaying varied distributions even within a short context window of a single time series. Clearly, frequency-level specialization is too coarse-grained to capture this level of diversity, underscoring the need for more fine-grained modeling approaches.

To address the aforementioned issues, this paper introduces $\textbf{\moiraimoe}$, an innovative solution for effective time series unified training, inspired by recent developments of Sparse Mixture of Experts (MoE) Transformers \citep{gshard, switch, deepseekmoe}. The core idea of $\moiraimoe$ is to utilize a single input/output projection layer while delegateing the modeling of diverse time series patterns to the sparse specialized experts in Transformer layers. With these designs, specialization of $\moiraimoe$ is achieved in a data-driven manner and operates at the token level. Moreover, this study investigates existing expert gating functions that generally use a randomly initialized linear layer for expert assignments \citep{shazeer2017outrageously, mixtralmoe} and introduces a new function that leverages cluster centroids derived from a pretrained model to guide expert allocations.

We extensively evaluate $\moiraimoe$ using a total of 39 datasets in in-distribution and zero-shot forecasting scenarios. The results confirm the superiority of $\moiraimoe$ over state-of-the-art foundation models including TimesFM \citep{timesfm}, Chronos \citep{chronos}, and $\moirai$ \citep{moirai}. Additionally, we conduct comprehensive model analyses, as the first attempt, to explore the inner workings of time series MoE foundation models. It reveals that $\moiraimoe$ acquires the capability to achieve frequency-invariant representations and essentially performs progressive denoising throughout the model. Our contributions are summarized as follows:
\begin{itemize}[leftmargin=*]
    \item We propose $\moiraimoe$, the first mixture-of-experts time series foundation models, achieving token-level model specialization in a data-driven manner. We introduce a new expert gating function for accurate expert assignments and improved performance.
    \item Extensive experiments on 39 datasets reveal that $\moiraimoe$ delivers up to 17\% performance improvements over $\moirai$ at the same level of model size, and outperforms other time series foundation models with up to 65$\times$ fewer activated parameters.
    \item We conduct thorough model analyses to deepen understanding of the inner workings of time series MoE foundation models and summarize valuable insights for future research.
\end{itemize}

\section{Related Work}
\paragraph{Foundation Models for Time Series Forecasting}
Time series foundation models~\citep{liang2024foundation} serve as versatile zero-shot forecasting tools. A key challenge in training these models is accommodating the high diversity of time series data, underscoring the possible need for designing specialization modules. Current approaches like TEMPO \citep{tempo} and UniTime \citep{unitime} utilize language-based prompts to identify data sources, facilitating model specialization at the dataset level. $\moirai$ \citep{moirai} advances this by focusing on a time series meta feature -- frequency. This method designs separate input/output projection layers for specific frequencies, allowing for frequency-specific specialization. Similarly, TimesFM \citep{timesfm} operates at this level of specialization by incorporating a frequency embedding dictionary to differentiate data. Some methods, like Chronos \citep{chronos}, Lag-LLaMA \citep{lag-llama}, Moment \citep{moment}, and Timer \citep{timer}, do not incorporate any specialization modules. Instead, they utilize the same architecture for all time series data, which can potentially increase the learning complexity and demand a large number of parameters to memorize the diverse input patterns. In this work, we propose to achieve automatic token-level specialization by using sparse mixture of experts, where diverse time series tokens are processed by specialized experts, while similar tokens share parameter space, thereby reducing learning complexity.

\paragraph{Sparse Mixture of Experts}
Mixture of experts (MoE) has emerged as an effective method for significantly scaling up model capacity while minimizing computation overhead in Large Language Models (LLMs) \citep{switch, deepseekmoe, llamamoe}. In this study, our motivation for using MoE is primarily centered on its capacity to enable token-level model specialization. A common approach for integrating MoE into Transformers involves replacing Feed-Forward Networks (FFNs) with MoE layers. An MoE layer consists of multiple expert networks and a gating function, where each expert shares the same structure as a standard FFN. The gating function is responsible for producing a gating vector that indicates the expert assignment. The assignment is usually sparse to maintain computational efficiency in the MoE layer, meaning that each token is generally processed by only one \citep{switch} or two \citep{deepspeedmoe, mixtralmoe} experts.

\section{Methodology}
\begin{figure}[t]
    \centering
    \includegraphics[width=0.98\textwidth]{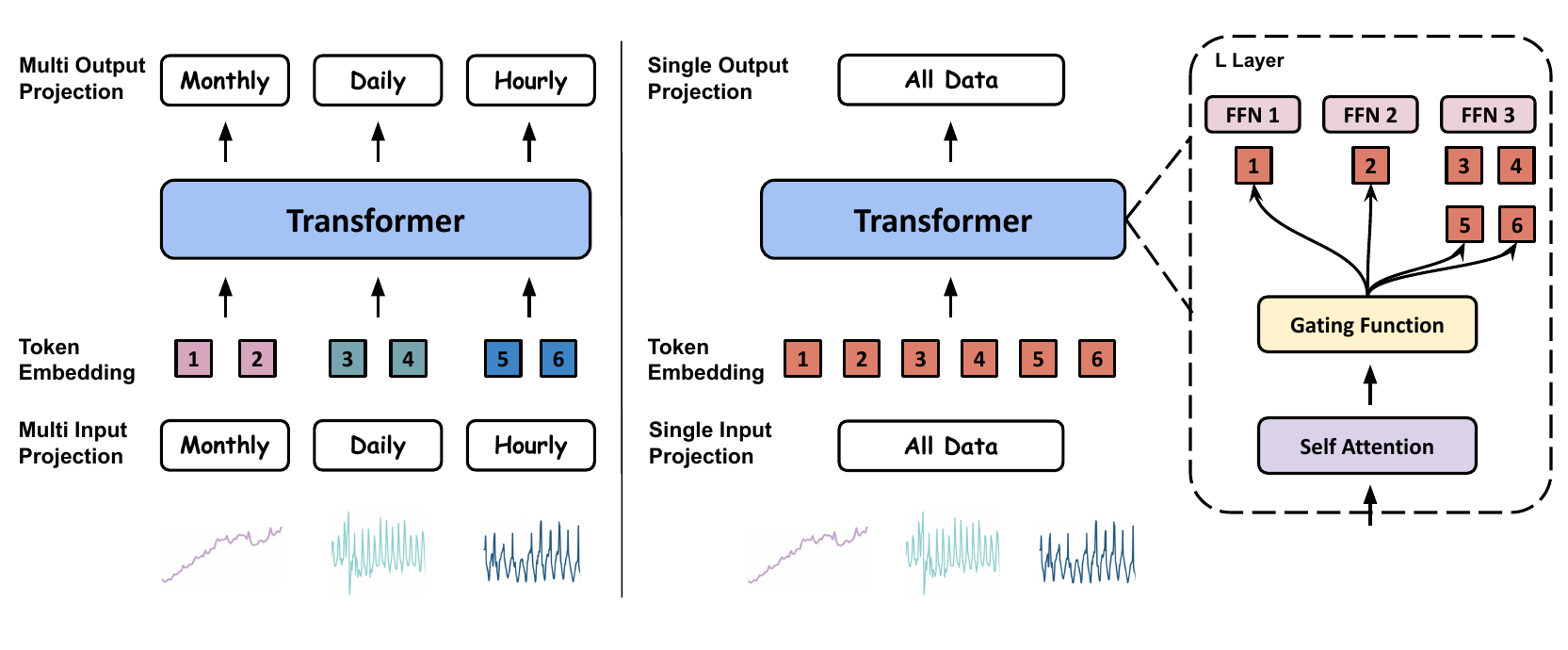}
    \caption{Comparison of $\moirai$ (left) and $\moiraimoe$ (right).}
    \label{fig:framework}
\end{figure}

In this section, we present $\moiraimoe$, a mixture-of-experts time series foundation model built upon $\moirai$ \citep{moirai}. Figure \ref{fig:framework} presents a comparison. While $\moiraimoe$ inherits many of the strengths of $\moirai$, its major enhancement lies in: rather than using multi heuristic-defined input/output projection layers to model time series with different frequencies, $\moiraimoe$ utilizes a single input/output projection layer while delegating the task of capturing diverse time series patterns to the sparse mixture of experts in the Transformer. In addition, $\moiraimoe$ proposes a novel gating function that leverages knowledge from a pretrained model, and adopts a decoder-only training objective to improve training efficiency by enabling parallel learning of various context lengths in a single model update. We describe each model component in the following parts.

\subsection{Time Series Token Construction}
Patching techniques, first introduced in PatchTST \citep{patchtst}, have become a prevalent method in many state-of-the-art time series models \citep{timesfm, unitime, moirai}. By aggregating adjacent time series data into patches, this technique effectively captures local semantic information and significantly reduces computational overhead when processing long inputs. Given a time series with length $S$, we segment it into non-overlapping patches of size $P$, resulting in a sequence of patches $\vx \in \mathbb{R}^{N \times P}$, where $N = \lceil \frac{S}{P} \rceil$.

We then normalize the patches to mitigate distribution shift issues \citep{nonstationary, timesnet}. In a decoder-only (autoregressive) model, where each patch predicts its succeeding patch, applying a causal normalizer to each patch is the most effective way to achieve accurate normalization. However, this approach generates $N$ subsequences with different lengths, diminishing the parallel training that decoder-only models typically offer. To address this, we introduce the masking ratio $r$ as a hyperparameter, which specifies the portion of the entire sequence used exclusively for robust normalizer calculation, without contributing to the prediction loss.

Finally, we forward the patches through a single projection layer to generate time series tokens $\vx \in \mathbb{R}^{N \times D}$, where $D$ is the dimension of the Transformers. And we pass on the capability of learning time series with diverse patterns to the vast number of parameters in the Transformer. This projection layer is implemented as a residual multi-layer perceptron to enhance representation capacity \citep{tide}.

\subsection{Mixture of Experts for Transformers}
A decoder-only Transformer \citep{llama3} is constructed by stacking $L$ layers of Transformer blocks, where each block can be represented as follows:
\begin{gather}
    \tilde{\vx}^{l-1} = \text{CSA}(\text{LN}(\vx^{l-1})) + \vx^{l-1} \\
    \vx^{l} = \text{FFN}(\text{LN}(\tilde{\vx}^{l-1})) + \tilde{\vx}^{l-1}
\end{gather}
where CSA, FFN, and LN denote a causal self-attention module, a feed-forward network, and the layer normalization, respectively. Following $\moirai$ \citep{moirai}, $\moiraimoe$ captures multivariate correlations by flattening all variates into a sequence. During causal attention, each token is allowed to attend to its preceding tokens, as well as preceding tokens from other variates.

Next, we establish the mixture of experts by replacing each FFN with a MoE layer, which is composed of $M$ expert networks $\{E_1, \ldots, E_M\}$ and a gating function $G$. Only a subset of experts is activated for each token, allowing experts to specialize in distinct patterns of time series data and ensuring computational efficiency. The output of the MoE layer is computed as:
\begin{gather}
    \sum^{M}_{i=1} G(\tilde{\vx}^{l-1})_i \cdot E_i(\tilde{\vx}^{l-1})
\end{gather}
where $E_i(\tilde{\vx}^{l-1})$ is the output of the $i$-th expert network, and $G(\tilde{\vx}^{l-1})_i$ is the $i$-th token-to-expert affinity score generated by the gating function. Following \cite{gshard, deepspeedmoe, mixtralmoe}, we set the number of activated experts to $K=2$.

\subsubsection{Gating Function}
\paragraph{Linear Projection as Gating Function.}
A popular and effective gating function takes the softmax over the TopK logits of a linear projection parameterized by $\mW_g \in \mathbb{R}^{D \times M}$ \citep{shazeer2017outrageously, mixtralmoe, deepseekmoe}:
\begin{gather}
    G(\tilde{\vx}^{l-1}) = \text{Softmax}(\text{TopK}(\tilde{\vx}^{l-1} \cdot \mW_g))
\end{gather}

However, the sparse gating can result in a load balancing issue \citep{shazeer2017outrageously}. To mitigate this, an auxiliary loss is typically introduced to encourage an even distribution of tokens across experts \citep{gshard, switch, mixtralmoe, deepseekmoe}. Formally, the load balancing loss for a batch $\mathcal{B}$ containing $T$ tokens is defined as:
\begin{gather}
    \mathcal{L}_{\text{load}} = M \sum^{M}_{i=1} \mathcal{D}_i \mathcal{P}_i, \text{where} \
    \mathcal{D}_i = \frac{1}{T} \sum_{\vx \in \mathcal{B}} \mathbb{1}\{\text{argmax} \  G(\tilde{\vx}^{l-1}) = i\}, 
    \mathcal{P}_i = \frac{1}{T} \sum_{\vx \in \mathcal{B}} G(\tilde{\vx}^{l-1})_i
\end{gather}
where $\mathcal{D}_i$ denotes the fraction of tokens routed to expert $i$ and $\mathcal{P}_i$ indicates the proportion of the gating probability allocated to expert $i$.

\paragraph{Token Clusters as Gating Function.}
In this work, we propose a new gating mechanism that leverages cluster centroids derived from the token representations of a pretrained model to guide expert allocations. The intuition behind this approach is that clusters of pretrained token embeddings more closely reflect the real distribution of the data, leading to more effective expert specialization compared to a randomly initialized linear projection layer. Specifically, we utilize the self-attention output representations $\tilde{\vx}^{l-1}$ of a pretrained model (in our case, we use the $\moirai$ model) and apply k-means clustering to generate clusters. The number of clusters is set to match the total number of experts. During MoE training, each token computes the Euclidean distance to each cluster centroid $\mC \in \mathbb{R}^{M \times D}$, and these distances serve as token-to-expert affinity scores for expert assignments:
\begin{gather}
    G(\tilde{\vx}^{l-1}) = \text{Softmax}(\text{TopK}(\text{Euclidean}(\tilde{\vx}^{l-1}, \mC)))
\end{gather}

\subsection{Training Objective}
Let $\vx_{t-l+1:t} = \{\vx_{t-l+1}, \ldots, \vx_{t}\}$ denote the context window of length $l$ for a token at position $t$. In this study, to facilitate both point and probabilistic forecasting, our goal is formulated as forecasting the predictive distribution of the next token $p(\vx_{t+1}|\phi)$ by predicting the mixture distribution parameters $\hat{\phi}$ \citep{moirai}. These parameters are derived from the output tokens of the Transformer, followed by a single output projection layer. The following negative log-likelihood is minimized during training:
\begin{gather}
     \mathcal{L}_{\text{pred}} = -\log p(\vx_{t+1} | \ \hat{\phi}), \ \hat{\phi} = f_{\vtheta}(\vx_{t-l+1:t})
\end{gather}

\begin{table}[b]
    \centering
    \caption{Model configurations of $\moirai$ and $\moiraimoe$.}
    \label{tab:model-size}%
    \resizebox{0.85\textwidth}{!}{
    \begin{tabular}{lccccccc}
        \toprule
        Model & Layers & $d$\textsubscript{model} &  $d$\textsubscript{ff} & Activated Params & Total Params & Activated Experts & Total Experts \\
        \midrule
        $\moiraismall$ & 6 & 384 & 1,024 & 14M & 14M & -- & -- \\
        $\moiraibase$ & 12 & 768 & 2,048 & 91M & 91M & -- & -- \\
        $\moirailarge$ & 24 & 1,024 & 2,736 & 310M & 310M & -- & -- \\
        $\moiraimoesmall$ & 6 & 384 & 512 & 11M & 117M & 2 & 32 \\
        $\moiraimoebase$ & 12 & 768 & 1,024 & 86M & 935M & 2 & 32 \\
        \bottomrule
    \end{tabular}%
    }
\end{table}%

\section{Experiments}
\subsection{$\moiraimoe$ Setup}
To ensure a fair comparison with $\moirai$ in terms of activated parameters, we configure the number of activated experts as $K = 2$ for $\moiraimoe$, resulting in 11M/86M activated parameters per token for $\moiraimoesmall$/$\moiraimoebase$, closely matching the dense model $\moiraismall$/$\moiraibase$ that contains 14M/91M activated parameters. The total number of experts $M$ is set to 32, yielding total parameter sizes of 117M for $\moiraimoesmall$ and 935M for $\moiraimoebase$. $\moiraimoelarge$ is not presented due to the significant requirements of computational resources. All $\moiraimoe$ models are trained on 16 A100 (40G) GPUs using a batch size of 1,024 and bfloat16 precision. The small and base model are trained for 50,000 and 250,000 steps on LOTSA \citep{moirai}, respectively. The patch size $P$ is set to 16 and the masking ratio $r$ for decoder-only training is 0.3 (the corresponding experiments are provided in Appendix \ref{sec:app-addition-res}). For optimization, we utilize the AdamW optimizer with lr = 1e-3, weight decay = 1e-1, $\beta_1$ = 0.9, $\beta_2$ = 0.98. We also apply a learning rate scheduler with linear warmup for the first 10,000 steps, followed by cosine annealing. The specific configurations are outlined in Table \ref{tab:model-size}.

\begin{figure}[t]
    \centering
    \includegraphics[width=\textwidth]{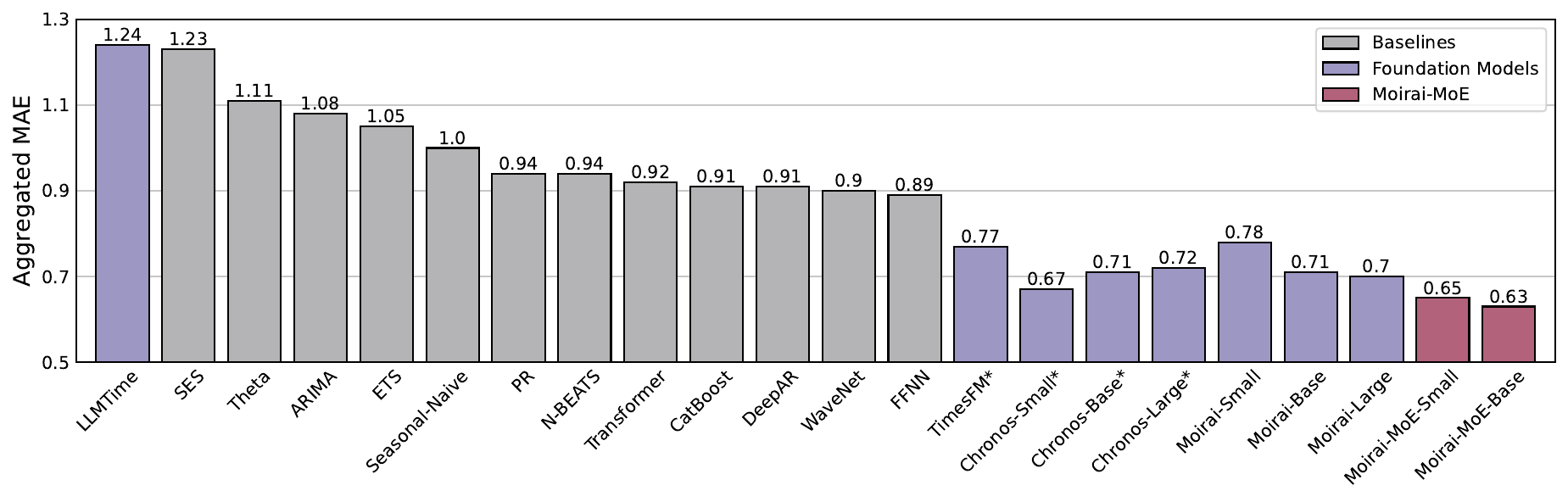}
    \caption{In-distribution forecasting evaluation using \textbf{29} datasets from Monash \citep{monash}. We use asterisks (*) to mark the methods that used the evaluation datasets here in their pretraining corpora. Aggregate MAE is reported, where the MAE for each dataset is normalized by the MAE of the seasonal naive forecast and the results are combined using the geometric mean.}
    \label{fig:in-distribution}
\end{figure}

\begin{table}[t]
    \centering
    \caption{Zero-shot performance of probabilistic and point forecasting. We use asterisks (\textbf{*}) to mark the non-zero-shot datasets because they were used in the pretraining corpus of TimesFM and Chronos. The Average column is normalized by seasonal naive, followed by geometric mean. Best average results are highlighted in \textcolor{red}{\textbf{red}}, and second best results are in \textcolor{blue}{\textbf{blue}}. Power: Turkey Power. Traffic: Istanbul Traffic. Weather: Jena Weather. BizITObs: BizITObs-L2C.}
    \label{tab:zero-shot}%
    \resizebox{\textwidth}{!}{
    \begin{tabular}{lccccccccccc>{\columncolor{gray!25}}c}
    \toprule
    \textbf{Method} & \textbf{Metric} & \textbf{Electricity} & \textbf{Solar} & \textbf{Power} & \textbf{ETT1} & \textbf{ETT2} & \textbf{Traffic} & \textbf{MDENSE} & \textbf{Walmart} & \textbf{Weather} & \textbf{BizITObs} & \textbf{Average} \\
    \midrule
    \multirow{2}[2]{*}{Seasonal Naive} & CRPS & 0.070 & 0.512 & 0.085 & 0.515 & 0.205 & 0.257 & 0.294 & 0.151 & 0.068 & 0.262 & 1.000 \\
    & MASE & 0.881 & 1.203 & 0.906 & 1.778 & 1.390 & 1.137 & 1.669 & 1.236 & 0.782 & 0.986 & 1.000 \\
    \midrule
    \multirow{2}[2]{*}{TiDE} & CRPS & 0.048 & 0.420 & 0.046 & 1.056 & 0.130 & 0.110 & 0.091 & 0.077 & 0.054 & 0.124 & 0.631 \\
    & MASE & 0.706 & 1.265 & 0.904 & 6.898 & 2.189 & 0.618 & 0.911 & 0.814 & 0.832 & 0.450 & 0.931 \\
    \midrule
    \multirow{2}[2]{*}{PatchTST} & CRPS & 0.052 & 0.518 & 0.054 & 0.304 & 0.131 & 0.112 & 0.070 & 0.082 & 0.059 & 0.074 & 0.549 \\
    & MASE & 0.753 & 1.607 & 1.234 & 1.680 & 2.168 & 0.653 & 0.732 & 0.867 & 0.844 & 0.266 & 0.808 \\
    \midrule
    \multirow{2}[2]{*}{iTransformer} & CRPS & 0.057 & 0.443 & 0.056 & 0.344 & 0.129 & 0.105 & 0.072 & 0.070 & 0.053 & 0.077 & 0.540 \\
    & MASE & 0.875 & 1.342 & 1.076 & 2.393 & 1.841 & 0.581 & 0.727 & 0.761 & 0.623 & 0.271 & 0.767 \\
    \midrule
    \multirow{2}[2]{*}{TimesFM} & CRPS & 0.045\textbf{*} & 0.456 & 0.037 & 0.280 & 0.113 & 0.131 & 0.070 & 0.067 & 0.042 & 0.080 & \textcolor{blue}{\textbf{0.488}} \\
    & MASE & 0.655\textbf{*} & 1.391 & 0.851 & 1.700 & 1.644 & 0.678 & 0.702 & 0.735 & 0.440 & 0.310 & 0.689 \\
    \midrule
    \multirow{2}[2]{*}{Chronos\textsubscript{S}} & CRPS & 0.043\textbf{*} & 0.389\textbf{*} & 0.038 & 0.360 & 0.097 & 0.124 & 0.087  & 0.079 & 0.089 & 0.087 & 0.543 \\
    & MASE & 0.629\textbf{*} & 1.193\textbf{*} & 0.717 & 1.799 & 1.431 & 0.622 & 0.834 & 0.849 & 0.606 & 0.301 & 0.694 \\
    \midrule
    \multirow{2}[2]{*}{Chronos\textsubscript{B}} & CRPS & 0.041\textbf{*} & 0.341\textbf{*} & 0.039 & 0.387 & 0.092 & 0.109 & 0.075 & 0.080 & 0.058 & 0.084 & 0.499 \\
    & MASE & 0.617\textbf{*} & 1.002\textbf{*} & 0.722 & 1.898 & 1.265 & 0.553 & 0.712 & 0.849 & 0.583 & 0.301 & \textcolor{blue}{\textbf{0.656}} \\
    \midrule
    \multirow{2}[2]{*}{Chronos\textsubscript{L}} & CRPS & 0.041\textbf{*} & 0.339\textbf{*} & 0.038 & 0.404 & 0.091 & 0.117 & 0.075 & 0.073 & 0.062 & 0.084 & 0.500 \\
    & MASE & 0.615\textbf{*} & 0.987\textbf{*} & 0.702 & 1.959 & 1.270 & 0.597 & 0.724 & 0.788 & 0.601 & 0.310 & 0.660 \\
    \midrule
    \multirow{2}[2]{*}{\moiraismall} & CRPS & 0.072 & 0.471 & 0.048 & 0.275 & 0.101 & 0.173 & 0.084 & 0.103 & 0.049 & 0.081 & 0.578 \\
    & MASE & 0.981 & 1.465 & 0.948 & 1.701 & 1.417 & 0.990 & 0.836 & 1.048 & 0.521 & 0.301 & 0.798 \\
    \midrule
    \multirow{2}[2]{*}{\moiraibase} & CRPS & 0.055 & 0.419 & 0.040 & 0.301 & 0.095 & 0.116 & 0.104 & 0.093 & 0.041 & 0.078 & 0.520 \\
    & MASE & 0.792 & 1.292 & 0.888 & 1.736 & 1.314 & 0.644 & 1.101 & 0.964 & 0.487 & 0.291 & 0.736 \\
    \midrule
    \multirow{2}[2]{*}{\moirailarge} & CRPS & 0.050 & 0.406 & 0.036 & 0.286 & 0.094 & 0.112 & 0.095 & 0.098 & 0.051 & 0.079 & 0.514 \\
    & MASE & 0.751 & 1.237 & 0.870 & 1.750 & 1.436 & 0.631 & 0.957 & 1.007 & 0.515 & 0.285 & 0.729 \\
    \midrule
    \multirow{2}[2]{*}{\moiraimoesmall} & CRPS & 0.046 & 0.429 & 0.036 & 0.288 & 0.093 & 0.108 & 0.071 & 0.090 & 0.056 & 0.081 & 0.497 \\
    & MASE & 0.719 & 1.222 & 0.737 & 1.750 & 1.248 & 0.563 & 0.746 & 0.927 & 0.476 & 0.298 & 0.670 \\
    \midrule
    \multirow{2}[2]{*}{\moiraimoebase} & CRPS & 0.041 & 0.382 & 0.034 & 0.296 & 0.091 & 0.100 & 0.071 & 0.088 & 0.057 & 0.079 & \textcolor{red}{\textbf{0.478}} \\
    & MASE & 0.638 & 1.161 & 0.725 & 1.748 & 1.247 & 0.510 & 0.721 & 0.918 & 0.509 & 0.290 & \textcolor{red}{\textbf{0.651}} \\
    \bottomrule
    \end{tabular}%
    }
\end{table}%

\subsection{Main Results}
\paragraph{In-distribution Forecasting.}
We begin with an in-distribution evaluation using a total of \textbf{29} datasets from the Monash benchmark \citep{monash}. Their training set are included in LOTSA \citep{moirai}, holding out the test set which we now use for assessments. Figure \ref{fig:in-distribution} summarizes the results based on the aggregated mean absolute error (MAE), in comparison with the baselines presented in the Monash benchmark and the recently released foundation models: TimesFM (200M) \citep{timesfm}, Chronos family \citep{chronos}: Chronos\textsubscript{S} (46M), Chronos\textsubscript{B} (200M), Chronos\textsubscript{L} (710M), and $\moirai$ family \citep{moirai}: $\moiraismall$ (14M), $\moiraibase$ (91M), $\moirailarge$ (310M). Full results are provided in Appendix \ref{sec:app-exp-details}. The evaluation results show that $\moiraimoe$ beats all competitors. In particular, $\moiraimoesmall$ drastically surpasses its dense counterpart $\moiraismall$ by 17\%, and also outperforms the larger models $\moiraibase$ and $\moirailarge$ by 8\% and 7\%, respectively. $\moiraimoebase$ delivers a further 3\% improvement over $\moiraimoesmall$. Compared to the foundation model Chronos, which $\moirai$ could not surpass, $\moiraimoe$ successfully bridges the gap and delivers superior results with up to 65$\times$ fewer activated parameters.

\paragraph{Zero-shot Forecasting.}
Next, we conduct an out-of-distribution evaluation on \textbf{10} datasets not included in LOTSA (see dataset details in Appendix \ref{sec:app-exp-details}). To establish a comprehensive comparison, we report results for both probabilistic and point forecasting, using continuous ranked probability score (CRPS) and mean absolute scaled error (MASE) as evaluation metrics. For baselines, we compare against foundation models TimesFM, Chronos, and $\moirai$, as well as state-of-the-art full-shot models trained on individual datasets: TiDE \citep{tide}, PatchTST \citep{patchtst}, and iTransformer \citep{itransformer}. The results are presented in Table \ref{tab:zero-shot}. \textbf{$\moiraimoebase$ achieves the best zero-shot performance, even outperforming TimesFM and Chronos, which include partial evaluation data in their pretraining corpora}. When compared to all sizes of $\moirai$, $\moiraimoesmall$ delivers a 3\%–14\% improvement in CRPS and an 8\%–16\% improvement in MASE. These improvements are remarkable, considering that $\moiraimoesmall$ has only 11M activated parameters -- 28$\times$ fewer than $\moirailarge$. 

\paragraph{Summary.}
Our extensive evaluation validates the effectiveness of $\moiraimoe$'s overall model design, demonstrates the strong generalization ability of $\moiraimoe$, and emphasizes the superiority of token-level specialization over frequency-level approaches (TimesFM, $\moirai$) and models without a specialization module (Chronos). $\moiraimoe$ also performs significantly better than the full-shot models trained separately on each dataset, highlighting the exceptional capabilities of the foundation model.

\subsection{Ablation Studies}

\begin{wraptable}{r}{0.5\textwidth} \vspace{-1.3em}
    \centering
    \caption{Model variants performance on Monash.}
    \label{tab:ablation}%
    \resizebox{0.5\textwidth}{!}{
    \begin{tabular}{lc}
    \toprule
    Model Variant & Aggregated MAE \\
    \midrule
    Multi Projection w/ Masked Encoder & 0.78 \\
    Multi Projection w/ Decoder-Only  & 0.75 \\
    Single Projection \& MoE w/ Decoder-Only & 0.65 \\
    \bottomrule
    \end{tabular}%
    }
\end{wraptable}%

\paragraph{Model Design.}
In the main results, we simultaneously enable the mixture of experts and switch the training objective from a masked encoder approach to a decoder-only approach. To ensure a more rigorous comparison, we conduct further experiments where only the learning objective is changed. Table \ref{tab:ablation} presents the Monash evaluation results using the small model, with the first and last rows representing $\moiraismall$ and $\moiraimoesmall$, respectively. This outcome suggests that altering the learning objective alone yields modest performance improvements, while the major gains stem from leveraging experts for automatic token-level specialization.

\begin{figure}[t]
    \centering
    \includegraphics[width=0.99\textwidth]{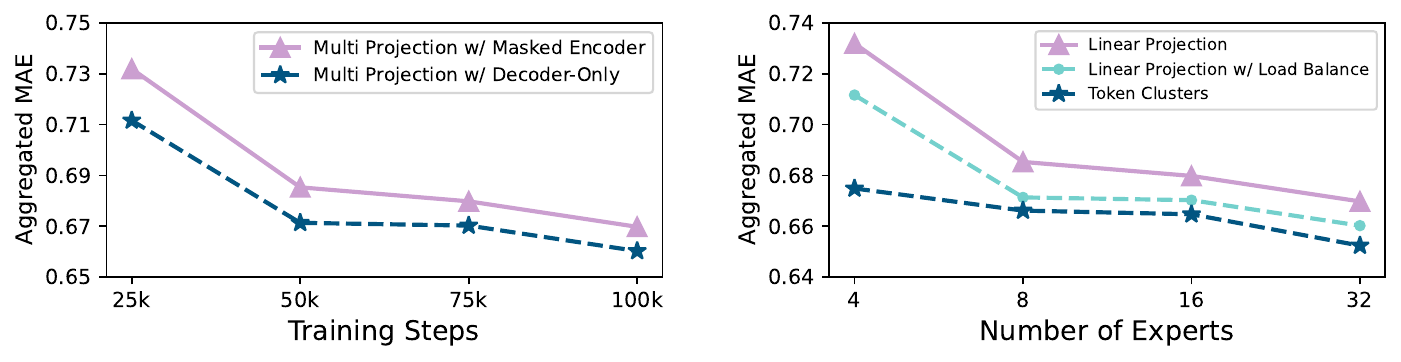}
    \caption{Ablation studies of the training objective and gating function using $\moiraimoesmall$.}
    \label{fig:ablation}
\end{figure}

\paragraph{Training Objective.}
We adopt the decoder-only training objective for its superior training efficiency compared to the masked encoder approach. To illustrate this, we conduct experiments with varying training steps, as shown in Figure \ref{fig:ablation} (left). The results show that the decoder-only approach consistently outperforms the masked encoder at each evaluated step. Moreover, decoder-only training with 50k steps achieves comparable performance to masked encoder training with 100k steps, highlighting the substantial efficiency gains provided by the decoder-only training objective.

\paragraph{Gating Function.}
In Figure \ref{fig:ablation} (right), we vary the total number of experts and examine the impact of different gating functions on performance. Across all gating functions, performance consistently improves as the number of experts increases. Notably, our proposed token clustering method proves to be consistently superior to the other gating function variants across all expert configurations. This indicates that the clustering approach aligns more closely with the inherent distribution of time series representations that have been optimized in pretraining, leading to more effective expert specialization compared to randomly learned-from-scratch gating. See more results in Appendix \ref{sec:app-expert-dist-gating}.

\begin{figure}[b]
    \centering
    \includegraphics[width=0.99\textwidth]{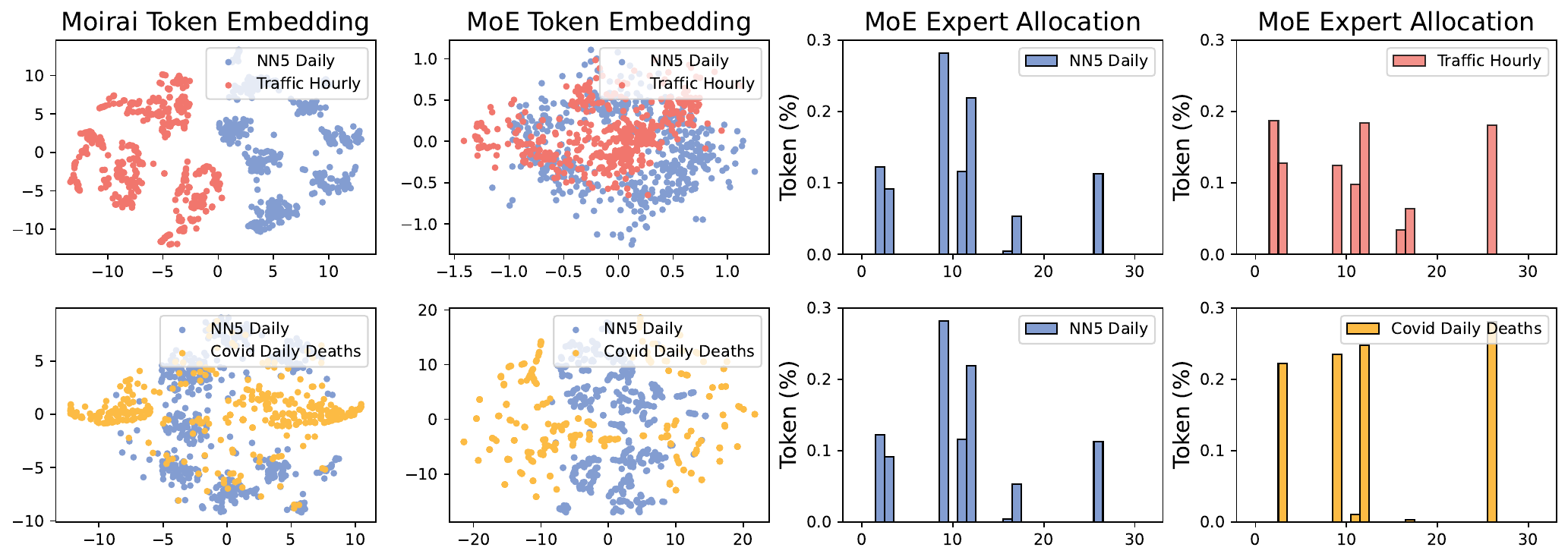}
    \caption{The first two columns are the comparison of embeddings from $\moiraismall$ and $\moiraimoesmall$. The last two columns are the expert assignment distributions of $\moiraimoesmall$ in layer 1: the x-axis corresponds to the 32 experts in a layer, and the y-axis is the proportion of tokens that choose experts.}
    \label{fig:case-study}
\end{figure}

\subsection{Model Analyses}
In this section, we delve deeper into the learned token embeddings and expert assignment distribution of $\moiraimoe$ to shed light on the inner workings of the time series MoE foundation model.

\paragraph{Obs 1: $\moiraimoe$ produces token embeddings in a data-driven way, effectively improving performance.}
In Figure \ref{fig:case-study}, we utilize the T-SNE visualization tool \citep{tsne} to compare the token embeddings generated from the input projection layers of $\moirai$ and $\moiraimoe$. (1) In the first row, we examine the NN5 Daily and Traffic Hourly datasets, which have different frequencies but exhibit similar underlying patterns (visualizations of these patterns can be found in Appendix \ref{sec:app-vis}). The figure illustrates that $\moirai$ produces distinct embeddings due to the use of separate frequency projection layers, while $\moiraimoe$ successfully blends their representations together. Their inherent similarities are further demonstrated by their comparable expert allocation distributions in the last two columns. (2) In the second row, we analyze another daily frequency dataset, Covid Daily Deaths, which shows distinct patterns compared to NN5 Daily. We observe that the embeddings of these two datasets overlap to some extent in the $\moirai$ model but are effectively separated in $\moiraimoe$. Furthermore, the Covid Daily dataset shows different expert selection choices than NN5 Daily due to different token embeddings. \textbf{The data-driven modeling paradigm of $\moiraimoe$ ultimately leads to significant performance boosts}, reducing the MAE of NN5 Daily from 5.37 to 4.04 (a 25\% improvement), the MAE of Traffic Hourly from 0.02 to 0.013 (a 35\% improvement), and the MAE of Covid  Daily Deaths from 124.32 to 119 (a 4\% improvement).

\begin{figure}[t]
    \centering
    \includegraphics[width=0.99\textwidth]{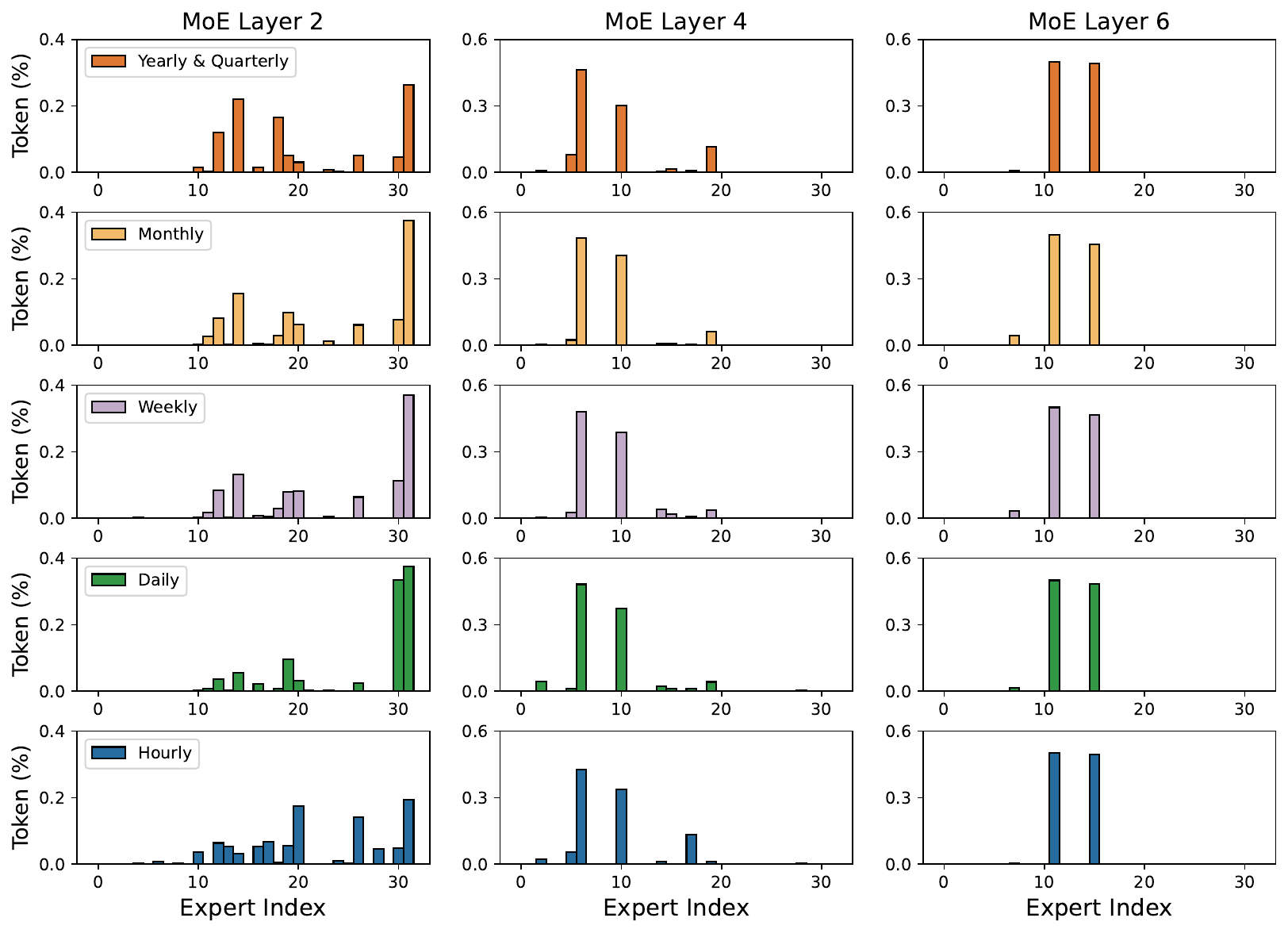}
    \caption{Visualization of the distribution of expert allocation for $\moiraimoesmall$ layers 2, 4, and 6 (the last layer) using the Monash benchmark grouped by time series frequency.}
    \label{fig:expert-freq}
\end{figure}

\paragraph{Obs 2: Different frequency data exhibit different expert selection distributions at shallow layers but similar distributions at deep layers.}
We present the expert allocation distributions on the Monash benchmark grouped by frequency in Figure \ref{fig:expert-freq}. In the shallow layers, expert selection is notably diverse, indicating that the model relies on multiple experts to manage the high level of short-term variability, such as cyclical, seasonal, or abrupt changes. As tokens are aggregated in deeper layers, the model shifts its focus to more generalizable temporal dependencies, such as broader trends and long-term patterns, that can be shared across different frequencies and leads to more concentrated experts being selected. By the final layer (layer 6), expert allocation becomes nearly identical across all frequencies, suggesting that the model has abstracted time series into high-level representations largely independent of the frequency. This evidence indicates that \textbf{$\moiraimoe$ effectively achieves frequency-invariant hidden representations}, which are crucial for model generalization \citep{cfa}. The shared parameter space in the last layer also shows that it is sufficient for generating representations needed to make diverse predictions.

\paragraph{Obs 3: Shallow layers have more routing preferences than deep layers.}
According to Figure \ref{fig:expert-freq}, as the layer index increases, expert selection gradually converges, with only 3 out of 32 experts being chosen by the final layer. This behavior contrasts with patterns observed in LLMs \citep{llamamoe}, where earlier layers typically concentrate on a limited number of experts to capture common linguistic features, while deeper layers target more task-specific characteristics. This divergence may stem from the dynamic and noisier nature of time series tokens, which are generated from small time windows, unlike language tokens derived from a fixed vocabulary. \textbf{Our findings suggest that denoising processes occur progressively throughout the model}. This observation aligns with conclusions from GPT4TS \citep{gpt4ts}, which found that as the layer depth increases, token vectors are projected into the low-dimensional top eigenvector space of input patterns. Additionally, we recognize that some experts in $\moiraimoe$ are rarely selected. Pruning these underutilized experts for model compression is left for future work.

\begin{figure}[t]
    \centering
    \includegraphics[width=0.99\textwidth]{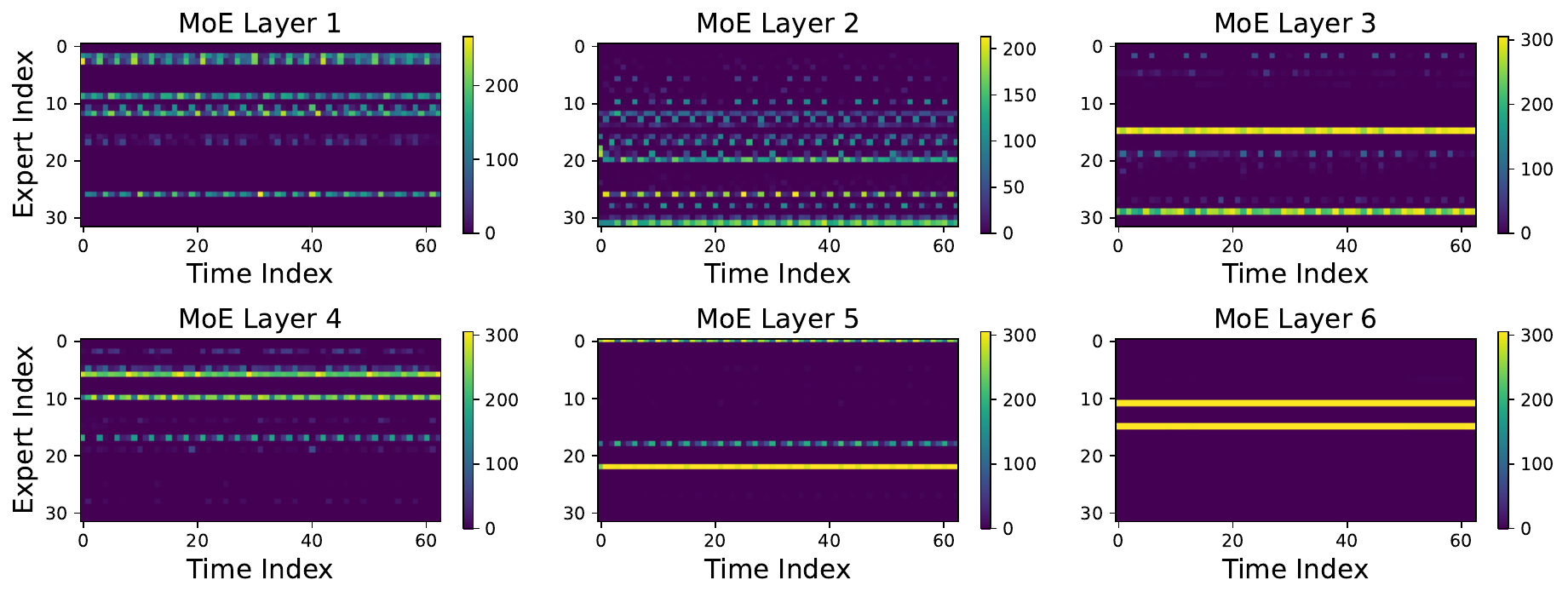}
    \caption{Visualization of expert allocation distributions for $\moiraimoesmall$. All MoE layers are presented. The x-axis is the time index of the 63 time series tokens, generated from 1,000 context lengths. The y-axis corresponds to the 32 experts in a layer.}
    \label{fig:expert-time-index}
\end{figure}

\paragraph{Obs 4: Expert allocation reflects time series periodicity patterns.}
To investigate the relationship between expert allocation and the positions of time series tokens, we use hourly data from the Monash repository with a minimum context length of 1,000 (e.g., the Traffic Hourly dataset). Figure \ref{fig:expert-time-index} visualizes the expert choices at each token position. In the shallow layers, we observe that expert selection follows periodic patterns, consistent with the actual patterns in the raw data, as shown in Figure \ref{fig:traffic-hourly}. This suggests that the model dynamically adapts to the cyclical nature of the traffic data, assigning specialized experts to manage tokens corresponding to distinct phases of the cycle, such as rising, peaks, and falling. In conclusion, \textbf{$\moiraimoe$ effectively learns to exploit time-based structures and the model specialization operates at the token level}.

\subsection{Efficiency Analyses}
In this section, we aim to validate whether the inference speeds of $\moirai$ and $\moiraimoe$ are comparable, as we have configured them with similar activated parameters. Additionally, due to the difference in the inference algorithms (the mask encoder in $\moirai$ predicts all tokens simultaneously, while the decoder-only approach in $\moiraimoe$ generates predictions autoregressively), we evaluate the inference cost on a subset of the Monash benchmark where the predicted token is one (corresponding to 16 time steps) to eliminate this discrepancy. To also compare to the foundation model Chronos, we set the context length to 512 and the number of sampling samples to 20, aligning with the settings used in Chronos.

We present the summarized results in Table \ref{tab:inference} and conclude that $\moiraimoesmall$ and $\moiraimoebase$ exhibit similar inference times to $\moiraismall$ and $\moiraibase$, respectively. These results highlight that $\moiraimoe$ not only maintains the same level of efficiency as $\moirai$ but also delivers substantial performance improvements. Additionally, when comparing $\moiraimoe$ to Chronos, which also employs autoregressive inference algorithms, we find that $\moiraimoe$ is significantly faster. This speed advantage stems from the fact that $\moiraimoe$ generates predictions using patches of size 16, while Chronos can be viewed as using a patch size of 1, which greatly affects its inference efficiency.

\begin{table}[h]
    \centering
    \caption{Inference cost evaluation. The values in brackets represent the parameter sizes of the foundation models. For MoE models, the two values indicate the number of activated parameters and the total number of parameters. The spent time is in seconds.}
    \label{tab:inference}%
    \resizebox{\textwidth}{!}{
    \begin{tabular}{lcccccccc}
    \toprule
    \multirow{2}[2]{*}{Model} & Chronos\textsubscript{S} & Chronos\textsubscript{B} & Chronos\textsubscript{L} & $\moiraismall$ & $\moiraibase$ & $\moirailarge$ & $\moiraimoesmall$ & $\moiraimoebase$ \\
    & (46M) & (200M) & (710M) & (14M) & (91M) & (310M) & (11M/117M) & (86M/935M) \\
    \midrule
    Spent Time (s) & 551 & 1,177 & 2,780 & 264 & 358 & 537 & 273 & 370 \\
    \bottomrule
    \end{tabular}%
    }
\end{table}%

\section{Conclusion}
In this work, we introduce the first time series MoE foundation model $\moiraimoe$ that utilizes sparse experts to model diverse time series patterns in a data-driven manner. Empirical experiments demonstrate that, by enabling automatic token-level specialization, $\moiraimoe$ not only achieves significant performance improvements over all sizes of its predecessor $\moirai$, but also outperforms other competitive foundation models like TimesFM and Chronos with much fewer activated parameters. Moreover, we conduct comprehensive model analyses to gain a deeper understanding of time series MoE foundation models.


\bibliography{iclr2025_conference}
\bibliographystyle{iclr2025_conference}

\clearpage

\appendix

\section{Experimental Details}
\label{sec:app-exp-details}
\paragraph{In-distribution Forecasting Datasets.}
Following $\moirai$ \citep{moirai}, we perform evaluations on 29 datasets from the Monash benchmark \citep{monash}, including M1 Monthly, M3 Monthly, M3 Other, M4 Monthly, M4 Weekly, M4 Daily, M4 Hourly, Tourism Quarterly, Tourism Monthly, CIF 2016, Australian Electricity Demand, Bitcoin, Pedestrian Counts, Vehicle Trips, KDD Cup 2018, Australia Weather, NN5 Daily, NN5 Weekly, Carparts, FRED-MD, Traffic Hourly, Traffic Weekly, Rideshare, Hospital, COVID Deaths, Temperature Rain, Sunspot, Saugeen River Flow, and US Births. The statistics of data are provided in Table \ref{tab:in-distribution-dataset}, and the full results of time series foundation models are shown in Table \ref{tab:in-distribution-full}.

\begin{table}[h]
    \centering
    \caption{Summary of datasets used in the in-distribution forecasting evaluations.}
    \label{tab:in-distribution-dataset}%
    \resizebox{0.7\textwidth}{!}{
    \begin{tabular}{lcccc}
    \toprule
    Dataset & Domain & Frequency & Number of Series & Prediction Length \\
    \midrule
    M1 Monthly & Econ/Fin & M & 617 & 18 \\
    M3 Monthly & Econ/Fin & M & 1,428 & 18 \\
    M3 Other & Econ/Fin & M & 174 & 8 \\
    M4 Monthly & Econ/Fin & M & 48,000 & 18 \\
    M4 Weekly & Econ/Fin & W & 359 & 13 \\
    M4 Daily & Econ/Fin & D & 4,227 & 14 \\
    M4 Hourly & Econ/Fin & H & 414 & 48 \\
    Tourism Quarterly & Econ/Fin & Q & 427 & 8 \\
    Tourism Monthly & Econ/Fin & M & 366 & 24 \\
    CIF 2016 & Econ/Fin & M & 72 & 12 \\
    Aus. Elec. Demand & Energy & 30T & 5 & 336 \\
    Bitcoin  & Econ/Fin & D & 18 & 30 \\
    Pedestrain Counts & Transport & H & 66 & 24 \\
    Vehicle Trips & Transport & D & 329 & 30 \\
    KDD Cup 2018 & Energy & H & 270 & 168 \\
    Australia Weather & Nature & D & 3,010 & 30 \\
    NN5 Daily & Econ/Fin & D & 111 & 56 \\
    NN5 Weekly & Econ/Fin & W & 111 & 8 \\
    Carparts & Sales & M & 2,674 & 12 \\
    FRED-MD & Econ/Fin & M & 107 & 12 \\
    Traffic Hourly & Transport & H & 862 & 168 \\
    Traffic Weekly & Transport & W & 862 & 8 \\
    Rideshare & Transport & H & 2,304 & 168 \\
    Hospital & Healthcare & M & 767 & 12 \\
    COVID Deaths & Healthcare & D & 266 & 30 \\
    Temperature Rain & Nature & D & 32,072 & 30 \\
    Sunspot & Nature & D & 1 & 30 \\
    Saugeen River Flow & Nature & D & 1 & 30 \\
    US Births & Healthcare & D & 1 & 30 \\
    \bottomrule
    \end{tabular}%
    }
\end{table}%

\begin{table}[h]
    \centering
    \caption{Full MAE results of time series foundation models on the Monash Benchmark. The other baseline results can be found in \citep{moirai}.}
    \label{tab:in-distribution-full}%
    \resizebox{\textwidth}{!}{
    \begin{tabular}{lcccccccccccccccccc}
    \toprule
    Dataset & Seasonal Naive & LLMTime & TimesFM & $\moirai$\textsubscript{Small} & $\moirai$\textsubscript{Base} & $\moirai$\textsubscript{Large} & Chronos\textsubscript{Small} & Chronos\textsubscript{Base} & Chronos\textsubscript{Large} & $\moiraimoe$\textsubscript{Small} & $\moiraimoe$\textsubscript{Base} \\
    \midrule
    M1 Monthly & 2,011.96 & 2,562.84 & 1,673.60 & 2,082.26 & 2,068.63 & 1,983.18 & 1,797.78 & 1,637.68 & 1,627.11 & 1,992.49 & 1,811.94 \\
    M3 Monthly & 788.95 & 877.97 & 653.57 & 713.41 & 658.17 & 664.03 & 644.38 & 622.27 & 619.79 & 646.07 & 617.31 \\
    M3 Other & 375.13 & 300.30 & 207.23 & 263.54 & 198.62 & 202.41 & 196.59 & 191.80 & 205.93 & 185.89 & 179.92 \\
    M4 Monthly & 700.24 & 728.27 & 580.20 & 597.60 & 592.09 & 584.36 & 592.85 & 598.46 & 584.78 & 569.25 & 544.08 \\
    M4 Weekly & 347.99 & 518.44 & 285.89 & 339.76 & 328.08 & 301.52 & 264.56 & 252.26 & 248.89 & 302.65 & 278.37 \\
    M4 Daily & 180.83 & 266.52 & 172.98 & 189.10 & 192.66 & 189.78 & 169.91 & 177.49 & 168.41 & 172.45 & 163.40 \\
    M4 Hourly & 353.86 & 576.06 & 196.20 & 268.04 & 209.87 & 197.79 & 214.18 & 230.70 & 201.14 & 241.58 & 217.35 \\
    Tourism Quarterly & 11,405.45 & 16,918.86 & 10,568.92 & 18,352.44 & 17,196.86 & 15,820.02 & 7,823.27 & 8,835.52 & 8,521.70 & 9,508.07 & 7,374.27 \\
    Tourism Monthly & 1,980.21 & 5,608.61 & 2,422.01 & 3,569.85 & 2,862.06 & 2,688.55 & 2,465.10 & 2,358.67 & 2,140.73 & 2,523.66 & 2,268.31 \\
    CIF 2016 & 743,512.31 & 599,313.84 & 819,922.44 & 655,888.58 & 539,222.03 & 695,156.92 & 649,110.99 & 604,088.54 & 728,981.15 & 453,631.21 & 568,283.48 \\
    Aus. Elec. Demand & 455.96 & 760.81 & 525.73 & 266.57 & 201.39 & 177.68 & 267.18 & 236.27 & 330.04 & 215.28 & 227.92 \\
    Bitcoin & 7.78E+17 & 1.74E+18 & 7.78E+17 & 1.76E+18 & 1.62E+18 & 1.87E+18 & 2.34E+18 & 2.27E+18 & 1.88E+18 & 1.55E+18 & 1.90E+18 \\
    Pedestrian Counts & 65.60 & 97.77 & 45.03 & 54.88 & 54.08 & 41.66 & 29.77 & 27.34 & 26.95 & 41.35 & 32.37 \\
    Vehicle Trips & 32.48 & 31.48 & 21.93 & 24.46 & 23.17 & 21.85 & 19.38 & 19.25 & 19.19 & 21.62 & 21.65 \\
    KDD Cup 2018 & 47.09 & 42.72 & 40.86 & 39.81 & 38.66 & 39.09 & 38.60 & 42.36 & 38.83 & 40.21 & 40.86 \\
    Australia Weather & 2.36 & 2.17 & 2.07 & 1.96 & 1.80 & 1.75 & 1.96 & 1.84 & 1.85 & 1.76 & 1.75 \\
    NN5 Daily & 8.26 & 7.10 & 3.85 & 5.37 & 4.26 & 3.77 & 3.83 & 3.67 & 3.53 & 4.04 & 3.49 \\
    NN5 Weekly & 16.71 & 15.76 & 15.09 & 15.07 & 16.42 & 15.30 & 15.03 & 15.12 & 15.09 & 15.74 & 15.29 \\
    Carparts & 0.67 & 0.44 & 0.50 & 0.53 & 0.47 & 0.49 & 0.52 & 0.54 & 0.53 & 0.45 & 0.44 \\
    FRED-MD & 5,385.53 & 2,804.64 & 2,237.63 & 2,568.48 & 2,679.29 & 2,792.55 & 938.46 & 1,036.67 & 863.99 & 1,651.76 & 2,273.61 \\
    Traffic Hourly & 0.013 & 0.030 & 0.009 & 0.020 & 0.020 & 0.010 & 0.013 & 0.012 & 0.010 & 0.013 & 0.014 \\
    Traffic Weekly & 1.19 & 1.15 & 1.06 & 1.17 & 1.14 & 1.13 & 1.14 & 1.12 & 1.12 & 1.13 & 1.14 \\
    Rideshare & 1.60 & 6.28 & 1.36 & 1.35 & 1.39 & 1.29 & 1.27 & 1.33 & 1.30 & 1.26 & 1.26 \\
    Hospital & 20.01 & 25.68 & 18.54 & 23.00 & 19.40 & 19.44 & 19.74 & 19.75 & 19.88 & 20.17 & 19.60 \\
    COVID Deaths & 353.71 & 653.31 & 623.47 & 124.32 & 126.11 & 117.11 & 207.47 & 118.26 & 190.01 & 119.00 & 102.92 \\
    Temperature Rain & 9.39 & 6.37 & 5.27 & 5.30 & 5.08 & 5.27 & 5.35 & 5.17 & 5.19 & 5.33 & 5.36 \\
    Sunspot & 3.93 & 5.07 & 1.07 & 0.11 & 0.08 & 0.13 & 0.20 & 2.45 & 3.45 & 0.10 & 0.08 \\
    Saugeen River Flow & 21.50 & 34.84 & 25.16 & 24.07 & 24.40 & 24.76 & 23.57 & 25.54 & 26.25 & 23.05 & 24.40 \\
    US Births & 1,152.67 & 1,374.99 & 461.58 & 872.51 & 624.30 & 476.50 & 432.14 & 420.08 & 432.14 & 411.61 & 385.24 \\
    \bottomrule
    \end{tabular}%
    }
\end{table}%

\clearpage

\paragraph{Zero-shot Forecasting Datasets.}
We conduct zero-shot evaluations on the datasets listed in Table \ref{tab:zero-shot-dataset}, which cover five domains and span frequencies ranging from minute-level to weekly. We use a non-overlapping rolling window approach, where the stride equals the prediction length. The test set consists of the last $h * r$ time steps, where $h$ is the forecast horizon and $r$ is the number of rolling evaluation windows. The validation set is defined as the last forecast horizon before the test set, while the training set includes all preceding data.

\begin{table}[h]
    \centering
    \caption{Summary of datasets used in the zero-shot forecasting evaluations.}
    \label{tab:zero-shot-dataset}%
    \resizebox{0.95\textwidth}{!}{
    \begin{tabular}{lcccc}
    \toprule
    Dataset & Domain & Frequency & Prediction Length & Rolling Evaluations \\
    \midrule
    Electricity \citep{trindade2015electricity} & Energy & H & 24 & 7 \\
    Solar \citep{lai2018modeling} & Energy & H & 24 & 7 \\
    Turkey Power \tablefootnote{https://www.kaggle.com/datasets/dharanikra/electrical-power-demand-in-turkey} & Energy & H & 24 & 7 \\
    ETT1 \citep{informer} & Energy & D & 30 & 3 \\
    ETT2 \citep{informer} & Energy & D & 30 & 3 \\
    Istanbul Traffic \tablefootnote{https://www.kaggle.com/datasets/leonardo00/istanbul-traffic-index} & Transport & H & 24 & 7 \\
    M-DENSE \citep{libcity} & Transport & D & 30 & 3 \\
    Walmart \citep{walmart2014sales} & Sales & W & 8 & 4 \\
    Jena Weather \citep{autoformer}  & Nature & 10T & 144 & 7 \\
    BizITObs-L2C \citep{automixer} & Web/CloudOps & 5T & 48 & 20 \\
    \bottomrule
    \end{tabular}%
    }
\end{table}%

\paragraph{Methods.} The following is a brief introduction to the models used in the evaluation process.
\begin{itemize}[leftmargin=*]
    \item TiDE \citep{tide} encodes the historical data of a time series along with covariates using dense multi-layer perceptrons (MLPs). It then decodes the time series while incorporating future covariates, also utilizing dense MLPs for this process.
    \item PatchTST \citep{patchtst} employs Transformer encoders combined with patching and channel independence techniques to enhance the performance of time series forecasting.
    \item iTransformer \citep{itransformer} treats independent time series as tokens to effectively capture multivariate correlations through self-attention. 
    \item LLMTime \citep{llmtime} is a method for time series forecasting that leverages Large Language Models by encoding numerical data as text and generating possible future values through text completions.
    \item TimesFM \citep{timesfm} is a decoder-only time series foundation model that pretrained on a large corpus of time series data, including both real-world and synthetic datasets.
    \item Chronos \citep{chronos} is an encoder-decoder time series foundation model that uses quantization to convert real numbers into discrete tokens.
    \item $\moirai$ \citep{moirai} is a time series foundation model trained on the LOTSA dataset, which contains over 27 billion observations across nine diverse domains.
    \item $\moiraimoe$ is proposed in this study, which is capable of achieving automatic token-level specialization.
\end{itemize}

\begin{table}[h]
    \centering
    \caption{Hyperparameter search values for TiDE, PatchTST, and iTransformer.}
    \label{tab:baseline-hyperparam}%
    \resizebox{0.4\textwidth}{!}{
    \begin{tabular}{lcc}
    \toprule
    & \textbf{Hyperparameter} & \textbf{Values} \\
    \midrule
    TiDE & hidden\_dim & \{64, 128, 256\} \\
         & num\_encoder\_layers & [2, 6] \\
         & num\_decoder\_layers & [2, 6] \\
    \midrule  
    PatchTST & d\_model & \{64, 128, 256\} \\
             & num\_encoder\_layers & [2,6] \\
    \midrule
    iTransformer & d\_model & \{128, 256, 512\} \\
                 & num\_encoder\_layers & [2, 4] \\
    \bottomrule
    \end{tabular}%
    }
\end{table}%

\paragraph{Hyperparameter Search.} For the three full-shot models used in zero-shot forecasting part, i.e., TiDE \citep{tide}, PatchTST \citep{patchtst}, and iTransformer \citep{itransformer}, we conduct hyperparameter search based on the values specified in Table \ref{tab:baseline-hyperparam}. In addition, we explore the learning rate in the range [1e-6, 1e-3] on a log scale, and set the context length as $l = m * h$, where $m$ is tuned in the range [2, 20], and $h$ is the prediction length. We implement a random search across these parameters over 15 training runs and report results based on the best validation CRPS.

\section{Additional Results}
\label{sec:app-addition-res}

\begin{figure}[h]
    \centering
    \includegraphics[width=0.99\textwidth]{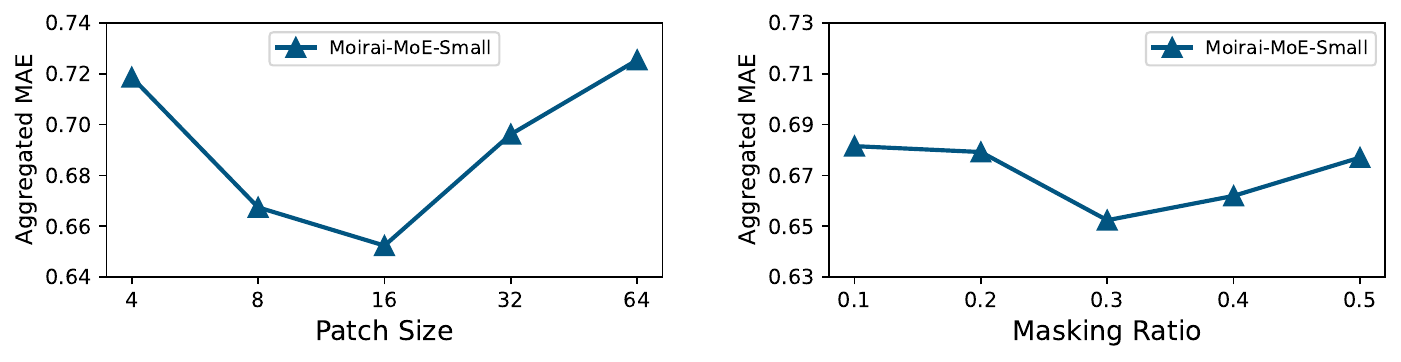}
    \caption{Effects of patch size and masking ratio using $\moiraimoesmall$.}
    \label{fig:hyperparam}
\end{figure}

\subsection{Effects of Patch Size}
In contrast to $\moirai$, which designs multiple input/output projection layers, each associated with a specific patch size, $\moiraimoe$ utilizes a single projection layer with a single patch size. In this part, we conduct experiments to examine the impact of different patch size choices. The evaluation results on the Monash benchmark are presented in Figure \ref{fig:hyperparam} (left), where the patch size of 16 yields the best performance. Increasing or decreasing this size results in performance degradation. Additionally, patch size affects inference speed; with a fixed context window, smaller patch sizes generate more time series tokens, increasing GPU memory usage and ultimately slowing down inference. For instance, using a patch size of 4 can take over a day to complete all evaluations. Our choice of a patch size of 16 not only delivers strong performance but also maintains a reasonable inference speed.

\subsection{Effects of Masking Ratio}
In this study, we introduce the masking ratio $r$ as a hyperparameter that determines the portion of the entire sequence used solely for robust normalizer calculation, helping to mitigate distribution shift issues. We conduct experiments to assess the effects of different masking ratios, with the evaluation results on the Monash benchmark shown in Figure \ref{fig:hyperparam} (right). A masking ratio of 0.3 delivers the best performance. A ratio of 0.1 uses too little data to compute a robust normalizer, potentially failing to accurately represent the overall sequence statistics. Conversely, a ratio of 0.5 masks half of the data, which may hinder the parallel learning efficiency in decoder-only training. Therefore, it is crucial to select an appropriate data range that is small enough to avoid excessive masking, yet sufficiently representative for robust normalizer computation.

\subsection{Expert Distributions of Different Gating Function}
\label{sec:app-expert-dist-gating}
In this part, we present an in-depth comparison of the different gating functions explored in this study.

First, we provide additional details on the implementation of the proposed token clustering method. The core idea of this approach is to leverage cluster centroids derived from the token representations of a pretrained model to guide expert allocations. Specifically, we perform inference on our training corpus, LOTSA, using data amount corresponding to 100 epochs. During this process, we extract the self-attention output representations from a pretrained $\moirai$ model and apply mini-batch k-means clustering to continuously update the clusters. The number of clusters is set to match the total number of experts. During the training of the MoE model, each token computes the Euclidean distance to each cluster centroid, and these distances are used as token-to-expert affinity scores for expert assignments. Empirical evaluations have demonstrated the effectiveness of this approach compared to randomly learned gating from scratch, indicating that the clustering method better aligns with the inherent distribution of time series representations.

\begin{figure}[t]
    \centering
    \includegraphics[width=0.99\textwidth]{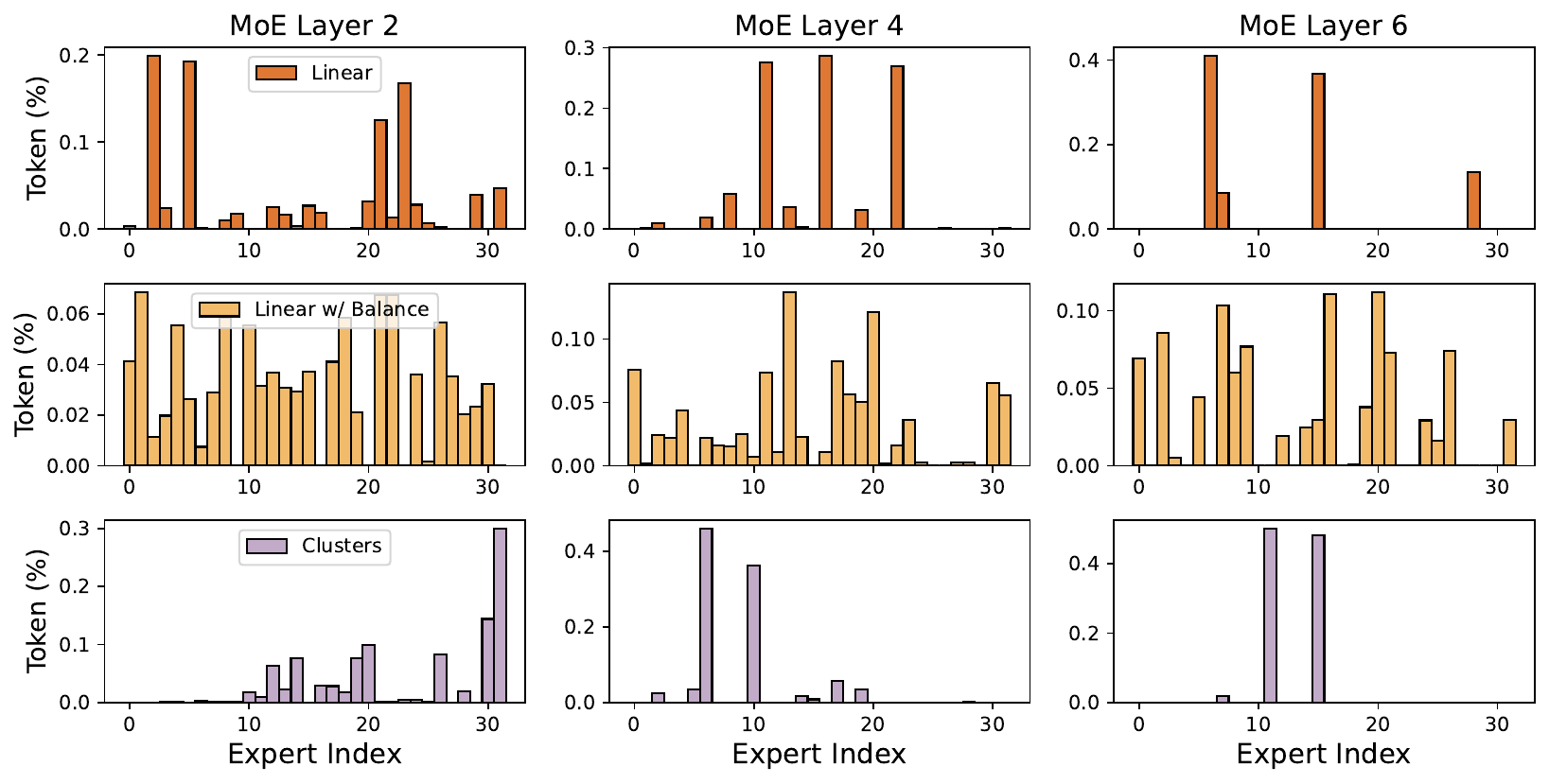}
    \caption{Visualization of the distribution of expert allocation for $\moiraimoesmall$ layers 2, 4, and 6 (the last layer) using all data from the Monash benchmark.}
    \label{fig:expert-gating}
\end{figure}

Using the three gating functions explored in this study, i.e., linear projection, linear projection with load balancing, and token clustering, we present their expert allocation distributions aggregated across all datasets in the Monash benchmark, as illustrated in Figure \ref{fig:expert-gating}. In terms of selection diversity, we observe the following relationships: Token Clusters (least diverse) $<$ Pure Linear Projection (neutral) $<$ Linear Projection with Load Balancing (most diverse). According to their performance results shown in Figure \ref{fig:ablation}, we can establish the following ranking: Token Clusters $>$ Linear Projection with Load Balancing $>$ Pure Linear Projection. Based on all these observations, we offer the following explanation:

\begin{itemize}[leftmargin=*]
    \item In the token clusters approach, the expert selections are less diverse because the routing is grounded in pretrained knowledge. The clustering step creates centroids that represent well-structured patterns in the data, and then tokens are routed to specific experts that are particularly suited to handle the type of data represented by their corresponding cluster. While this targeted routing reduces diversity, it enhances performance due to the selection of experts based on more meaningful criteria.
    \item The addition of load balancing loss increases the diversity of expert selection by spreading the workload and encouraging the use of all experts more evenly. This diversity prevents over-reliance on specific experts, potentially improving generalization and performance compared to pure linear projection. However, this approach might be less targeted than clustering, since it still depends on a learned gating function rather than pretrained centroids.
    \item  In the pure linear projection method, the gating function is entirely learned from scratch. Without any additional constraints (like load balancing), certain experts might get selected more often than others, leading to a neutral level of diversity. Since there is no mechanism to encourage exploration (like load balancing) or specialized routing (like clustering), performance remains lower than the other methods.
\end{itemize}

\section{Visualization}
\label{sec:app-vis}
\begin{figure}[h]
    \centering
    \includegraphics[width=0.99\textwidth]{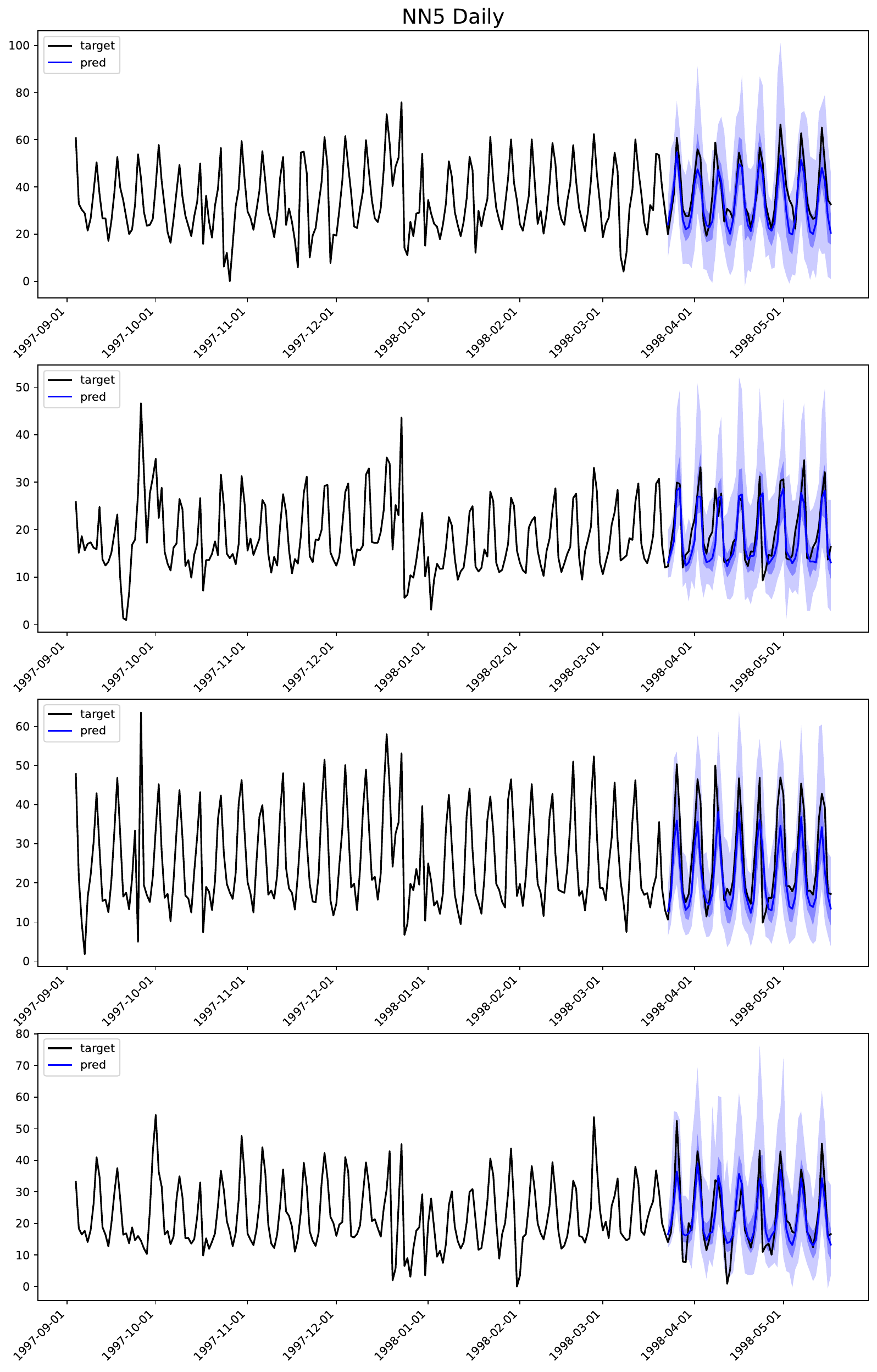}
    \caption{Visualization of NN5 Daily data, including both context length and forecast results.}
    \label{fig:nn5-daily}
\end{figure}

\begin{figure}[h]
    \centering
    \includegraphics[width=0.99\textwidth]{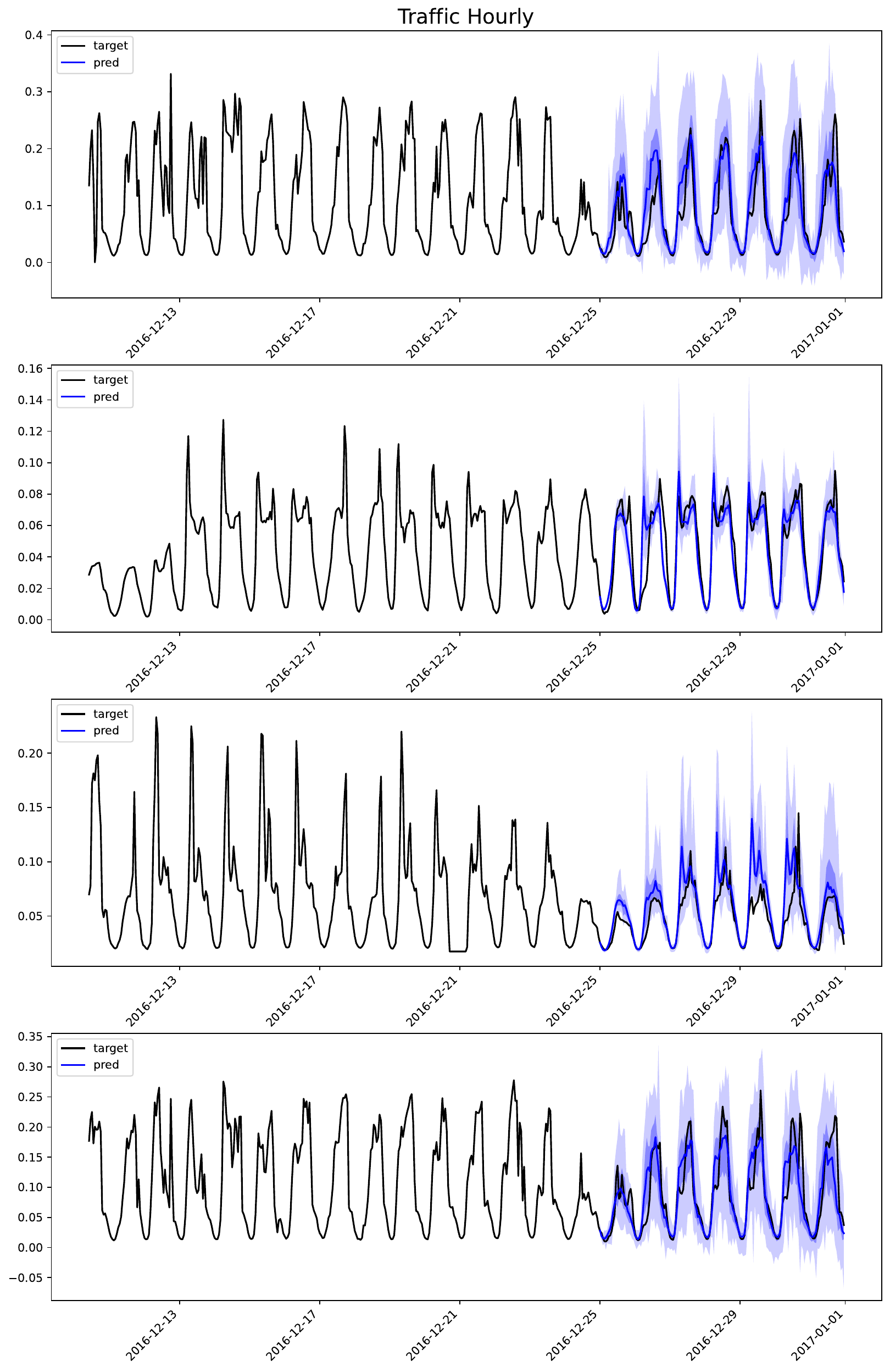}
    \caption{Visualization of Traffic Hourly data, including both context length and forecast results.}
    \label{fig:traffic-hourly}
\end{figure}

\begin{figure}[h]
    \centering
    \includegraphics[width=0.99\textwidth]{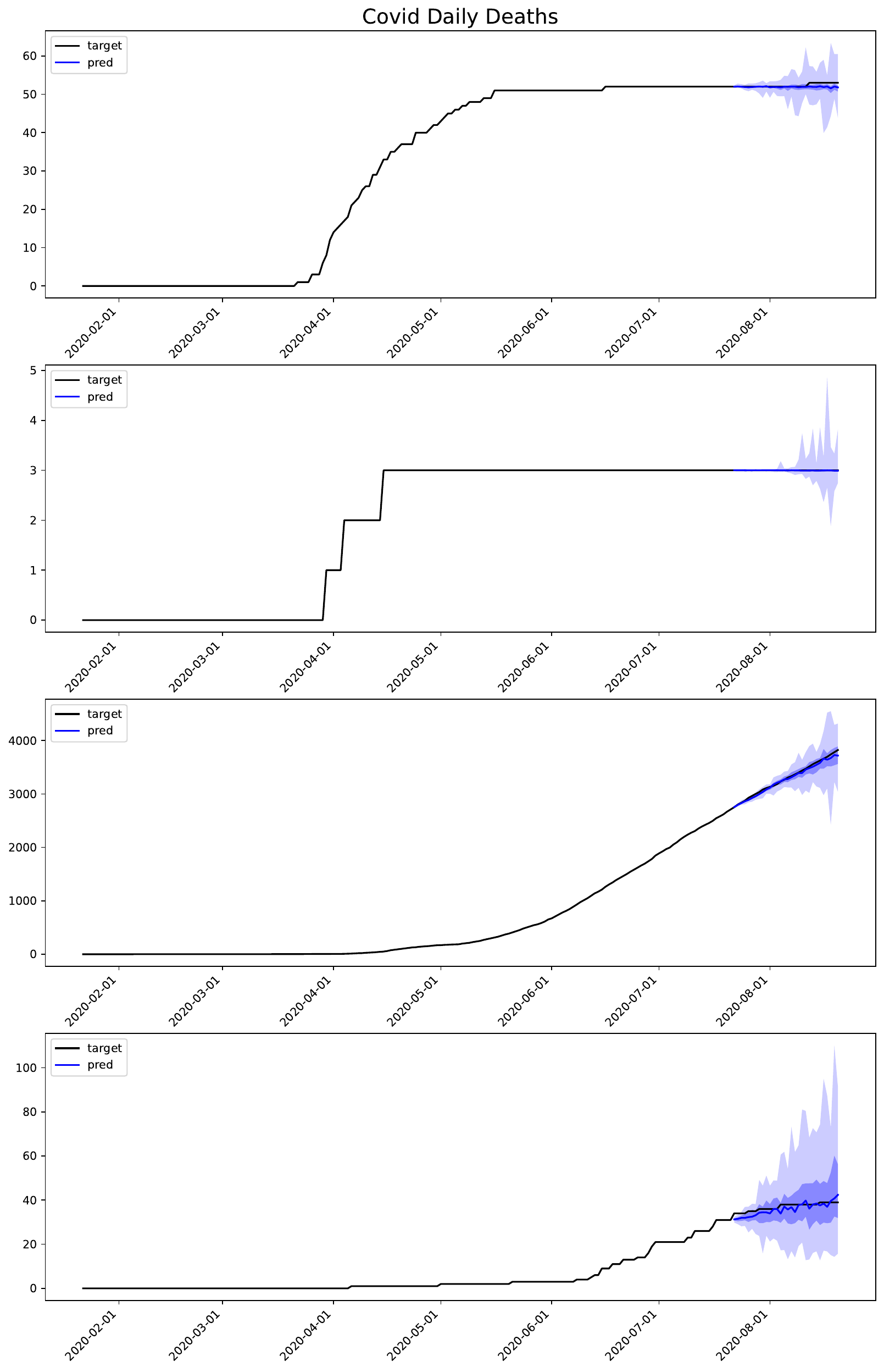}
    \caption{Visualization of Covid Daily Deaths, including both context length and forecast results.}
    \label{fig:covid-deaths}
\end{figure}

In this section, we visualize the datasets used in the model analyses (NN5 Daily (Figure \ref{fig:nn5-daily}), Traffic Hourly (Figure \ref{fig:traffic-hourly}), and Covid Daily Deaths (Figure \ref{fig:covid-deaths})) to facilitate understanding of the patterns within the time series data.

\end{document}